\documentclass[letterpaper,10pt,conference]{ieeeconf}

\IEEEoverridecommandlockouts		
\makeatletter
\let\NAT@parse\undefined
\makeatother

\makeatletter
\def\input@path{{sections/}}
\makeatother

\usepackage{cite}
\usepackage{amsmath,amssymb,amsfonts}
\usepackage{algorithmic}
\usepackage{graphicx}
\usepackage{textcomp}
\usepackage{xcolor}
\def\BibTeX{{\rm B\kern-.05em{\sc i\kern-.025em b}\kern-.08em
		T\kern-.1667em\lower.7ex\hbox{E}\kern-.125emX}}

\usepackage[left=0.75in, right=0.75in, bottom=0.75in, top=0.75in]{geometry}
\usepackage{comment}

\graphicspath{ {/figures/} }
\usepackage{float}					
\usepackage{tikz}                   
\usepackage{tikzscale}              
\usetikzlibrary{calc}               
\usepackage[caption=false,font=footnotesize]{subfig}	
\usepackage{pgfplots}               
\pgfplotsset{compat=newest}         
\pgfplotsset{plot coordinates/math parser=false}    
\newlength\fwidth                   
\definecolor{mycolor1}{rgb}{0.00000,0.44700,0.74100}    
\definecolor{mycolor2}{rgb}{0.85000,0.32500,0.09800}    
\definecolor{mycolor3}{rgb}{0.92900,0.69400,0.12500}    

\usepackage{mathtools}				
\DeclareMathOperator{\relu}{ReLU}	

\usepackage{amsthm}
\theoremstyle{plain}
\newtheorem{theorem}{Theorem}
\newtheorem{proposition}{Proposition}
\newtheorem{corollary}{Corollary}
\newtheorem{lemma}{Lemma}
\theoremstyle{definition}
\newtheorem{definition}{Definition}

\theoremstyle{remark}
\newtheorem{remark}{Remark}

\usepackage{hyperref}		
\hypersetup{%
	colorlinks=true,%
	linkcolor=black,%
	urlcolor=black,%
	citecolor=black,%
	filecolor=black%
}							
\usepackage[all]{hypcap}	

\title{\Large \bf
Tightened Convex Relaxations for Neural Network Robustness Certification
}

\author{%
	Brendon G.\ Anderson, Ziye Ma, Jingqi Li, and Somayeh Sojoudi%
	\thanks{%
	    The authors are with the University of California, Berkeley. Somayeh Sojoudi is also with the Tsinghua-Berkeley Shenzhen Institute. Emails: \texttt{\{\href{mailto:bganderson@berkeley.edu}{bganderson},\href{mailto:ziyema@berkeley.edu}{ziyema},\href{mailto:jingqili@berkeley.edu}{jingqili},\href{mailto:sojoudi@berkeley.edu}{sojoudi}\}@berkeley.edu}.}%
	\thanks{%
		This work was supported by grants from AFOSR, ONR, and NSF.}%
}

\begin{document}
\maketitle
\thispagestyle{empty}
\pagestyle{empty}

\begin{abstract}
In this paper, we consider the problem of certifying the robustness of neural networks to perturbed and adversarial input data. Such certification is imperative for the application of neural networks in safety-critical decision-making and control systems. Certification techniques using convex optimization have been proposed, but they often suffer from relaxation errors that void the certificate. Our work exploits the structure of ReLU networks to improve relaxation errors through a novel partition-based certification procedure. The proposed method is proven to tighten existing linear programming relaxations, and asymptotically achieves zero relaxation error as the partition is made finer. We develop a finite partition that attains zero relaxation error and use the result to derive a tractable partitioning scheme that minimizes the worst-case relaxation error. Experiments using real data show that the partitioning procedure is able to issue robustness certificates in cases where prior methods fail. Consequently, partition-based certification procedures are found to provide an intuitive, effective, and theoretically justified method for tightening existing convex relaxation techniques.
\end{abstract}

\section{Introduction}
\label{sec: introduction}
Recent successes of neural networks can be found in nearly all forms of data-driven decision making problems. In particular, both classical and modern problems within control theory have been addressed using neural networks, e.g., control of nonlinear systems \cite{kumpati1990identification,lewis1998neural}, data-driven system identification \cite{liu2012nonlinear,bansal2016learning,yeung2019learning}, and adaptive and self-learning control \cite{nguyen1990neural,johnson2001neural}. With their increasing prevalence, neural networks have begun to find applications in highly sensitive data-driven decision-making problems involving the control of safety-critical systems, such as autonomous vehicles \cite{bojarski2016end,wu2017squeezedet} and the power grid \cite{kong2017short,muralitharan2018neural,pan2019deepopf}. The common underlying principle among these systems is that decisions and control actions must be robust against fluctuations in the measurements or inputs to the decision making algorithm. As a result, much effort has been placed on developing methods to certify the robustness of neural networks to perturbations in their input data \cite{wong2017provable,raghunathan2018semidefinite,xiang2018reachability,weng2018towards,zhang2018efficient,fazlyab2019safety,royo2019fast,jin2019power}. Due to the vast range of network architectures, their inherent nonconvexity, and computational burdens arising with large-scale networks, the development of efficient and reliable certification methods remains an ongoing effort.

A common deterministic certification procedure is to verify that all possible unknown inputs are mapped to outputs that the network operator classifies as safe \cite{wong2017provable,royo2019fast}. From this perspective, certification amounts to proving that the image of an input uncertainty set is contained within a prescribed safe set. Such a worst-case analysis is naturally formulated as a robust optimization (RO) problem \cite{ben2009robust,bertsimas2011theory}, however, approaching the problem from a general RO framework neglects the informative structure of the network architecture. Furthermore, even when the input uncertainty set is convex, its output set may be nonconvex, which renders the certification an NP-complete nonconvex optimization problem \cite{katz2017reluplex,weng2018towards}. To overcome these issues, researchers have proposed various relaxations to over-approximate the output set of the network by a convex one and perform the certification on the easier-to-analyze convex set. One of the simplest and most popular approximation classes is based on linear program (LP) relaxations \cite{wong2017provable}.

In the case that a convex outer approximation of the original nonconvex output set is contained in the safe set, a certificate of robustness for the true network can be obtained. An immediate problem arises with these convex relaxations: if the convex outer approximation of the output set is too loose, the relaxation may not issue a certificate even in the case the true network is robust. To tighten the outer approximation, more sophisticated and computationally demanding convex relaxations have been proposed in the literature, such as the semidefinite programming and quadratically-constrained semidefinite programming techniques \cite{raghunathan2018semidefinite,fazlyab2019safety}.

\subsection{Partition-Based Certification}
In this paper, we focus on feedforward ReLU networks, which are popular due to their simplicity, fast training speeds, and non-vanishing gradient property\cite{xiang2018reachability}. Our approach to certifying these networks is based on partitioning the input uncertainty set and solving simple linear programs over each input part. Partitioning heuristics have been applied in areas such as robust optimization \cite{bertsimas2016multistage} and deep learning \cite{montufar2014number}, and are often found to tighten bounds on optimization objectives. Furthermore, partitioning naturally allows for parallelization of the optimization, resulting in computational advantages over centralized methods. Our proposed method is closely related to solving mixed integer programming (MIP) problems, as ReLU robustness certification can be expressed as an MIP. However, mixed integer reformulations of ReLU constraints introduce new integer variables for each neuron in the network, making MIP approaches, such as the outer approximation algorithm in \cite{duran1986outer}, unnecessarily large in dimension. Instead, our partition-based approach directly encodes the integral nature of the ReLU constraints without adding extra variables.

Previous works that apply partitioning to network certification include \cite{xiang2018reachability}, where the authors perform a reachability analysis for the safety verification of neural network controllers. However, that method is restricted to hyperrectangular partitions of both the input uncertainty set and the resulting outer approximations. The authors of \cite{royo2019fast} use duality arguments to propose a novel partitioning scheme; however, the designed algorithm only considers splitting box-shaped uncertainty sets in half along coordinate axes. Not only are the current partition-based methods too restrictive in their partition structure and accordingly produce unnecessarily loose outer approximations, but they also lack mathematical support for the effectiveness of the partitioning in tightening the relaxations.

\subsection{Contributions}
In an effort to improve relaxation errors, we exploit the nature of ReLU networks to achieve the following goals:
\begin{enumerate}
    \item Prove that partitioning tightens existing linear program relaxations, and define the notion of Lipschitz relaxations to show that relaxation error converges to zero as partitions become finer;
    \item Show that an intelligently designed finite partition attains zero relaxation error, and use this insight to derive a computationally tractable partitioning scheme that minimizes worst-case relaxation error;
    \item Demonstrate on real data that the optimal partitioning scheme sufficiently reduces relaxation error to certify robustness where prior methods fail.
\end{enumerate}
The contributions of this paper culminate into a theoretically justified and empirically validated robustness certification procedure that combines simple and efficient linear program models, computational parallelizability, and optimal relaxation tightening.

\subsection{Organization}
In Section \ref{sec: notations}, we define some mathematical notations. Section \ref{sec: problem_statement} introduces the robustness certification problem and linear program relaxation. In Section \ref{sec: properties_of_partitioned_relaxations}, we introduce the notion of partitioning and analyze its properties when applied to the robustness certification of ReLU networks. In Section \ref{sec: optimal_partitioning}, we further develop the theory to study the optimality of partitions and propose an optimal partitioning strategy. We provide illustrative examples in Section \ref{sec: simulation_results} and conclude in Section \ref{sec: conclusions}.

\section{Notations}
\label{sec: notations}
We write the sets of $n$-vectors and $m\times n$ matrices with real-valued elements as $\mathbb{R}^n$ and $\mathbb{R}^{m\times n}$, respectively. For $X,Y\in\mathbb{R}^{m\times n}$, we write $X\le Y$ to mean $X_{ij}\le Y_{ij}$ for all $i\in\{1,2,\dots,m\}$ and all $j\in\{1,2,\dots,n\}$. We write the Hadamard (element-wise) product between $X$ and $Y$ as $X\odot Y$ and the Hadamard division of $X$ by $Y$ as $X\oslash Y$. Furthermore, for $f\colon \mathbb{R}\to\mathbb{R}$, we define $f(X)$ to be an $m\times n$ matrix whose $(i,j)$ element is equal to $f(X_{ij})$ for all $i\in\{1,2,\dots,m\}$ and all $j\in\{1,2,\dots,n\}$. In particular, let the ReLU function be denoted as $\relu(\cdot) = \max\{0,\cdot\}$.

\section{Problem Statement}
\label{sec: problem_statement}
\subsection{Network Description}
Consider a $K$-layer ReLU neural network defined by
\begin{equation}
\begin{aligned}
x^{[0]} ={}& x, \\
\hat{z}^{[k]} ={}& W^{[k-1]}x^{[k-1]}+b^{[k-1]}, \\
x^{[k]} ={}& \relu(\hat{z}^{[k]}), \\
z ={}& x^{[K]},
\end{aligned} \label{eq: network_description}
\end{equation}
for all $k\in\{1,2,\dots,K\}$, where $x\in\mathbb{R}^{n_x}$ is the input to the neural network, $z\in\mathbb{R}^{n_z}$ is the output, and $\hat{z}^{[k]}\in\mathbb{R}^{n_k}$ is the preactivation of the $k^{\text{th}}$ layer. The parameters $W^{[k]}\in\mathbb{R}^{n_{k+1}\times n_k}$ and $b^{[k]}\in\mathbb{R}^{n_{k+1}}$ are the weight matrix and bias vector applied to the $k^{\text{th}}$ layer's activation $x^{[k]}\in\mathbb{R}^{n_k}$, respectively. Without loss of generality, assume that the bias terms are accounted for in the activations $x^{[k]}$, thereby setting $b^{[k]}=0$ for all layers $k$. Let the function $f\colon \mathbb{R}^{n_x}\to\mathbb{R}^{n_z}$ denote the map $x\mapsto z$ defined by (\ref{eq: network_description}).

\subsection{Input Uncertainty, Relaxed Network Constraint, and Safe Sets}
We consider the scenario in which the network inputs are unknown but contained in a compact set $\mathcal{X}\subseteq\mathbb{R}^{n_x}$. We call $\mathcal{X}$ the \emph{input uncertainty set}, which is assumed to be a convex polytope. In the literature of neural network robustness certification, the input uncertainty set is commonly modeled as $\mathcal{X} = \{x\in\mathbb{R}^{n_x} : \|x - \bar{x}\|_\infty \le \epsilon\}$, where $\bar{x}\in\mathbb{R}^{n_x}$ is a nominal input to the network and $\epsilon>0$ \cite{wong2017provable,raghunathan2018semidefinite}.

The bounds on the input, as defined by $\mathcal{X}$, implicitly define bounds on the preactivation at each layer. That is, $x\in\mathcal{X}$ implies that there exist bounds $l^{[k]},u^{[k]}\in\mathbb{R}^{n_k}$ such that $l^{[k]}\le \hat{z}^{[k]}\le u^{[k]}$ for all $k\in\{1,2,\dots,K\}$. Although one can create an outer approximation of these bounds, we consider the true bounds $l^{[k]}$ and $u^{[k]}$ as tight, i.e., $z^{[k]}=l^{[k]}$ for some $x\in\mathcal{X}$ and similarly for the upper bound $u^{[k]}$. From these bounds, we relax the $k^{\text{th}}$ ReLU constraint in (\ref{eq: network_description}) to its convex envelope, which leads to a relaxed ReLU constraint set associated with the $k^{\text{th}}$ layer:
\begin{equation}
\begin{aligned}
\mathcal{N}^{[k]} ={}& \{(x^{[k-1]},x^{[k]})\in\mathbb{R}^{n_{k-1}}\times\mathbb{R}^{n_k} : \\
&{} x^{[k]} \le u^{[k]} \odot (\hat{z}^{[k]}-l^{[k]}) \oslash (u^{[k]}-l^{[k]}), \\
&{} x^{[k]} \ge 0, ~ x^{[k]} \ge \hat{z}^{[k]}, ~ \hat{z}^{[k]}=W^{[k-1]}x^{[k-1]}\}.
\end{aligned} \label{eq: relu_constraint_set}
\end{equation}
Define the \emph{relaxed network constraint set} as
\begin{equation}
\begin{aligned}
\mathcal{N} ={}& \{(x,z)\in\mathbb{R}^{n_x}\times \mathbb{R}^{n_z} : (x,x^{[1]})\in\mathcal{N}^{[1]}, \\
&{} (x^{[1]},x^{[2]})\in\mathcal{N}^{[2]},\dots,(x^{[K-1]},z)\in\mathcal{N}^{[K]}\}.
\end{aligned} \label{eq: network_constraint_set}
\end{equation}
In essence, $\mathcal{N}$ is the set of all feasible input-output pairs of the network satisfying the relaxed ReLU constraint at each layer. Since the bounds $l^{[k]}$ and $u^{[k]}$ are determined by the input uncertainty set $\mathcal{X}$, the set $\mathcal{N}^{[k]}$ is also determined by $\mathcal{X}$ for all layers $k$.

\begin{figure}[ht]
    \centering
    \includegraphics[width=0.5\linewidth]{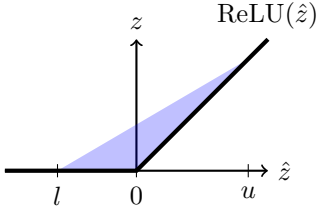}
    \caption{Relaxed ReLU constraint set at a single neuron for a one-layer network. The convex envelope $\mathcal{N}$ is shaded.}
    \label{fig: relaxed_relu_constraint}
    \vspace*{-\baselineskip}
\end{figure}

\begin{remark}
	In the context of one-layer networks (i.e., $K=1$), the single relaxed ReLU constraint set coincides with the relaxed network constraint set: $\mathcal{N}^{[1]}=\mathcal{N}$. Therefore, for $K=1$ we drop the $k$-notation from $z$, $\hat{z}$, $x$, $W$, $l$, $u$, and $\mathcal{N}$. A visualization of $\mathcal{N}$ is given in Fig.\ \ref{fig: relaxed_relu_constraint} for this case.
\end{remark}

\begin{remark}
	Consider a one-layer ReLU constraint relaxed according to (\ref{eq: relu_constraint_set}). Suppose that $l< u<0$. A simple calculation shows that $\mathcal{N} = \{(x,0)\in\mathbb{R}^{n_x}\times\mathbb{R}^{n_z} : Wx = l\}$ due to the inequalities $l\leq Wx\leq u$. That is, the set of input-output pairs that are feasible for the relaxed network constraints exclude many possible inputs that are feasible for the input uncertainty constraints. The same problem occurs when $0<l< u$. To overcome this issue, we impose the conditions that $l\le 0\le u$ and $l<u$ so that the certification procedure considers all possible inputs in $\mathcal{X}$.
\end{remark}

Now, consider a set $\mathcal{S}\subseteq\mathbb{R}^{n_z}$, termed the \emph{safe set}. As is common in the adversarial machine learning literature, we consider (possibly unbounded) polyhedral safe sets defined as the intersection of a finite number of half-spaces: $\mathcal{S} = \{z\in\mathbb{R}^{n_z} : C z \le d\}$, where $C\in\mathbb{R}^{n_\mathcal{S}\times n_z}$ and $d\in\mathbb{R}^{n_\mathcal{S}}$ are given. An output $z \in \mathcal{S}$ is said to be $\emph{safe}$.

\subsection{Robustness Certification}
The goal is to certify that all inputs in $\mathcal{X}$ map to safe outputs in $\mathcal{S}$. If this is successfully accomplished, the network is said to be \emph{certifiably robust}. Formally, this certificate is written as $f(\mathcal{X})\subseteq \mathcal{S}$, or equivalently
\begin{equation*}
\sup_{x\in\mathcal{X}}c_i^\top f(x) \le d_i ~ \text{for all $i\in\{1,2,\dots,n_{\mathcal{S}}\}$},
\end{equation*}
where $c_i^\top$ is the $i^{\text{th}}$ row of $C$. Thus, the certification procedure amounts to solving an optimization problem corresponding to each $c_i$. In the sequel, we focus on a single optimization problem, namely $\sup_{x\in\mathcal{X}}c^\top f(x)$, since the generalization to the case $n_\mathcal{S}>1$ is straightforward. With no loss of generality, assume that $d=0$ (if $d\ne 0$, one can first solve the optimization for $d=0$ and then shift the corresponding result). Note that the proposed mathematical framework encapsulates the popular certification that a classification network will not misclassify any adversarial inputs within a bounded uncertainty set.

\begin{figure}[ht]
    \centering
    \includegraphics[width=0.65\linewidth]{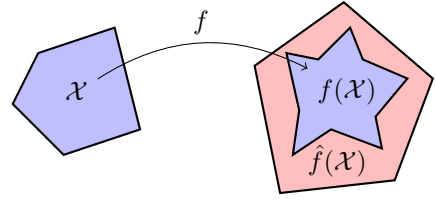}
    \caption{The convex outer approximation of the nonconvex set $f(\mathcal{X})$ is $\hat{f}(\mathcal{X})$. If the outer approximation is safe, i.e., $\hat{f}(\mathcal{X})\subseteq \mathcal{S}$, then so is $f(\mathcal{X})$.}
    \label{fig: outer_approximation}
    \vspace*{-\baselineskip}
\end{figure}

In general, the optimization $\sup_{x\in\mathcal{X}}c^\top f(x)$ is a nonconvex problem and $f(\mathcal{X})$ is a nonconvex set, and therefore computing a robustness certificate is intractable. To circumvent this issue, one can instead certify that a convex outer approximation of $f(\mathcal{X})$ is safe, as this inherently certifies the safety of the true nonconvex set $f(\mathcal{X})$, and hence certifies the robustness of the network. This process is illustrated in Fig.\ \ref{fig: outer_approximation}.

The robustness certification problem can be written as
\begin{equation}
\begin{aligned}
f^*(\mathcal{X}) ={}& \sup\{c^\top z : z=f(x), ~ x\in\mathcal{X} \}.
\end{aligned} \label{eq: robustness_certification_problem}
\end{equation}
The nonconvexity of (\ref{eq: robustness_certification_problem}) comes from the nonlinear equality constraint $z=f(x)$. Note that for all $x\in\mathcal{X}$, the equality $z=f(x)$ implies that $(x,z)\in\mathcal{N}$. Therefore, to avoid the nonconvex equality constraint, one can use the relaxed network constraint set to solve the following surrogate LP relaxation \cite{wong2017provable}:
\begin{equation}
\begin{aligned}
\hat{f}^*(\mathcal{X}) ={}& \sup\{c^\top z : (x,z)\in\mathcal{N}, ~ x\in\mathcal{X} \}.
\end{aligned} \label{eq: lp_relaxation}
\end{equation}
The suprema in (\ref{eq: robustness_certification_problem}) and (\ref{eq: lp_relaxation}) are assumed to be attained.

Due to the relaxation introduced in (\ref{eq: lp_relaxation}), it holds that
\begin{equation}
\begin{aligned}
f^*(\mathcal{X})\le{}& \hat{f}^*(\mathcal{X}).
\end{aligned} \label{eq: relaxation_bound}
\end{equation}
Therefore, a sufficient condition for the network to be certifiably robust is that $\hat{f}^*(\mathcal{X})\le 0$. In the case $\hat{f}^*(\mathcal{X}) > 0$, the relaxation cannot certify whether or not the true network is robust, since it may still hold that $f^*(\mathcal{X})\le 0$. The remainder of this paper is dedicated to optimally tightening the bound (\ref{eq: relaxation_bound}) while maintaining the advantageous convexity and computational properties of the LP relaxation.

\section{Properties of Partitioned Relaxations}
\label{sec: properties_of_partitioned_relaxations}
In this section, we investigate the notion of input partitioning and rigorously derive guarantees on the effectiveness of partitioned relaxations for the robustness certification problem. We will first show that by partitioning the input uncertainty set and solving separate LP relaxations over each part, a useful upper bound for the unrelaxed problem (\ref{eq: robustness_certification_problem}) can be obtained. In particular, the partitioning method yields a valid relaxation of (\ref{eq: robustness_certification_problem}).

\subsection{Validation of Partitioned Relaxations}

\begin{definition}[Partition]
	\label{def: partition}
	The collection $\{\mathcal{X}^{(j)} \subseteq \mathcal{X} : j\in\{1,2,\dots,p\}\}$ is said to be a \emph{partition} of the input uncertainty set $\mathcal{X}$ if $\mathcal{X}=\cup_{j=1}^p\mathcal{X}^{(j)}$ and $\mathcal{X}^{(j)}\cap \mathcal{X}^{(k)} = \emptyset$ for all $j \ne k$. The set $\mathcal{X}^{(j)}$ is called the $j^{\text{th}}$ \emph{input part}.
\end{definition}

\begin{proposition}[Partitioned relaxation bound]
	\label{prop: partitioned_relaxation_bound}
	Let $\{\mathcal{X}^{(j)} \subseteq \mathcal{X} : j\in\{1,2,\dots,p\}\}$ be a partition of $\mathcal{X}$. Then, it holds that
	\begin{equation}
	f^*(\mathcal{X}) \le \max_{j\in\{1,2,\dots,p\}}\hat{f}^*(\mathcal{X}^{(j)}). \label{eq: partitioned_relaxation_bound}
	\end{equation}
\end{proposition}

\begin{proof}
	Assume that $f^*(\mathcal{X}) > \max_{j\in\{1,2,\dots,p\}}\hat{f}^*(\mathcal{X}^{(j)})$. Then,
	\begin{equation}
	f^*(\mathcal{X})>\hat{f}^*(\mathcal{X}^{(j)}) ~ \text{for all $j\in\{1,2,\dots,p\}$}. \label{eq: partitioned_relaxation_bound_assumption}
	\end{equation}
	Let $(x^*,z^*)$ denote an optimal solution to the unrelaxed problem (\ref{eq: robustness_certification_problem}), i.e., $x^*\in\mathcal{X}$, $z^*=f(x^*)$, and
	\begin{equation}
	c^\top z^* = f^*(\mathcal{X}). \label{eq: partitioned_relaxation_bound_optimality}
	\end{equation}
	Since $\cup_{j=1}^p\mathcal{X}^{(j)} = \mathcal{X}$, there exists $j^*\in\{1,2,\dots,p\}$ such that $x^*\in\mathcal{X}^{(j^*)}$. Since $x^*\in\mathcal{X}^{(j^*)}$ and $z^*=f(x^*)$, it holds that $(x^*,z^*)\in\mathcal{N}^{(j^*)}$, where $\mathcal{N}^{(j^*)}$ is the relaxed network constraint set defined by $\mathcal{X}^{(j^*)}$. Therefore, 
	\begin{multline*}
	c^\top z^* \le \sup\{c^\top z : x\in\mathcal{X}^{(j^*)}, ~ (x,z)\in\mathcal{N}^{(j^*)}\} \\
	= \hat{f}^*(\mathcal{X}^{(j^*)}) < f^*(\mathcal{X}),
	\end{multline*}
	where the first inequality comes from the feasibility of $(x^*,z^*)$ over the $j^{*\text{th}}$ subproblem and the final inequality is due to (\ref{eq: partitioned_relaxation_bound_assumption}). This contradicts the optimality of $(x^*,z^*)$ given in (\ref{eq: partitioned_relaxation_bound_optimality}). Hence, (\ref{eq: partitioned_relaxation_bound}) must hold.
\end{proof}

\subsection{Tightening of the LP Relaxation}

The objective is to show that by partitioning the input uncertainty set, the linear program relaxation bound in (\ref{eq: relaxation_bound}) is improved. The result will be presented for one-layer networks for simplicity, but the conclusion naturally generalizes to multi-layer networks.

\begin{proposition}[Improving the relaxation bound]
	\label{prop: improving_the_relaxation_bound}
	Consider a one-layer feedforward neural network. Let $\{\mathcal{X}^{(j)} \subseteq \mathcal{X} : j\in\{1,2,\dots,p\}\}$ be a partition of $\mathcal{X}$. For the $j^{\text{th}}$ input part $\mathcal{X}^{(j)}$, denote the corresponding preactivation bounds by $l^{(j)}$ and $u^{(j)}$, where $l \le l^{(j)} \le Wx \le u^{(j)} \le u$ for all $x\in\mathcal{X}^{(j)}$. Then, it holds that
	\begin{equation}
	\max_{j\in\{1,2,\dots,p\}} \hat{f}^*(\mathcal{X}^{(j)}) \le \hat{f}^*(\mathcal{X}). \label{eq: improving_the_relaxation_bound}
	\end{equation}
\end{proposition}

\begin{proof}
	Let $j\in\{1,2,\dots,p\}$. It will be shown that $\mathcal{N}^{(j)} \subseteq \mathcal{N}$. Let $(x,z)\in\mathcal{N}^{(j)}$. Define $u'=u^{(j)}$, $l'=l^{(j)}$, and
	\begin{align*}
	g(x) ={}& u\odot(Wx-l)\oslash(u-l), \\
	g'(x) ={}& u'\odot (Wx-l')\oslash(u'-l').
	\end{align*}
	Then, by letting $\Delta g(x) = g(x) - g'(x) = a\odot (Wx) + b$, where
	\begin{align*}
	a ={}& u\oslash(u-l) - u'\oslash(u'-l'), \\
	b ={}& u'\odot l'\oslash(u'-l') - u\odot l\oslash(u-l),
	\end{align*}
	the following relations are derived for all $i\in\{1,2,\dots,n_z\}$:
	\begin{multline*}
	g^*_i \coloneqq \inf_{\{x : l'\le Wx\le u'\}} (\Delta g(x))_i \ge \inf_{\{\hat{z} : l'\le \hat{z}\le u'\}} (a\odot \hat{z} + b)_i \\
	= \inf_{\{\hat{z}_i : l_i' \le \hat{z}_i \le u'_i\}} (a_i \hat{z}_i + b_i) =
	\begin{aligned}
	\begin{cases}
	a_i l_i' + b_i & \text{if $a_i\ge 0$}, \\
	a_i u_i' + b_i & \text{if $a_i< 0$}.
	\end{cases}
	\end{aligned}
	\end{multline*}
	In the case that $a_i \ge 0$, we have that
	\begin{equation*}
	g_i^* \ge a_i l_i' + b_i = \frac{u_i}{u_i-l_i}(l_i'-l_i) \ge 0,
	\end{equation*}
	where the final inequality comes from the fact that $u\ge 0$, $l'\ge l$, and $u> l$. On the other hand, if $a_i < 0$, it holds that
	\begin{equation*}
	g_i^* \ge a_i u_i' + b_i = \frac{u_i'-u_i}{u_i-l_i}l_i \ge 0,
	\end{equation*}
	where the final inequality comes from the fact that $u'\le u$, $l\le 0$, and $u>l$. Therefore, $g^* = (g^*_1,g^*_2,\dots,g^*_{n_z})\ge 0$, which implies that $\Delta g(x) = g(x) - g'(x) \ge 0$ for all $x$ such that $l^{(j)}=l'\le Wx\le u'=u^{(j)}$. Hence, since $(x,z)\in\mathcal{N}^{(j)}$, it holds that $z\ge 0$, $z\ge Wx$, and
	\begin{equation*}
	z \le g'(x) \le g(x) = u\odot(Wx-l)\oslash(u-l).
	\end{equation*}
	Therefore, we have that $(x,z)\in\mathcal{N}$.
	
	Since $\mathcal{X}^{(j)}\subseteq \mathcal{X}$ (by definition) and $\mathcal{N}^{(j)}\subseteq \mathcal{N}$, it holds that the solution to the problem over the smaller feasible set gives a lower bound to the original solution: $\hat{f}^*(\mathcal{X}^{(j)}) \le \hat{f}^*(\mathcal{X})$. Finally, since $j$ was chosen arbitrarily, this implies the desired inequality (\ref{eq: improving_the_relaxation_bound}).
\end{proof}

\subsection{Asymptotic Exactness of Partitioned Relaxations}
In this section, we define the notion of Lipschitz continuity of a relaxation. We use this property to show that, under appropriate conditions, LP relaxations asymptotically approach the true problem as the partition becomes finer.

\begin{definition}[$L$-Lipschitz relaxation]
	\label{def: lipschitz_relaxation}
	A neural network $f$ is said to have an \emph{$L$-Lipschitz continuous relaxation} (with respect to $\mathcal{N}$ on $\mathcal{X}$) if there exists a finite constant $L\in\mathbb{R}$ such that
	\begin{equation*}
	|c^\top z_1^* - c^\top z_2^*|\le L\|x_1-x_2\|_2 ~ \text{for all $x_1,x_2\in\mathcal{X}$},
	\end{equation*}
	where $c^\top z_i^* = \sup\{c^\top z : (x_i,z)\in\mathcal{N}\}$ for all $i\in\{1,2\}$.
\end{definition}

\begin{remark}
	In the case the relaxed network constraint set is exact (i.e., $(x^{[k-1]},x^{[k]})\in\mathcal{N}^{[k]}$ if and only if $x^{[k]}=\relu(W^{[k-1]}x^{[k-1]})$ for all layers $k\in\{1,2,\dots,K\}$), the relation $x_i\in \mathcal{X}$ implies that $c^\top z_i^* = c^\top f(x_i)$, and so the $L$-Lipschitz continuity of the relaxation reduces to the classical $L$-Lipschitz continuity of the function $c^\top f$ over the set $\mathcal{X}$.
\end{remark}

\begin{lemma}[Lipschitz LP relaxations]
	\label{lem: lipschitz_lp_relaxations}
	All $K$-layer neural networks defined by (\ref{eq: network_description}) have $L$-Lipschitz continuous LP relaxations.
\end{lemma}

\begin{proof}
	Let $x_1,x_2\in\mathcal{X}$. By Theorem 2.4 in \cite{mangasarian1987lipschitz}, there exists a finite constant $\beta\in\mathbb{R}$ such that for all $z_1^*\in\arg\max\{ c^\top z:(x_1,z)\in\mathcal{N} \}$ there exists $z_2^*\in\arg\max\{c^\top z:(x_2,z)\in\mathcal{N}\}$ satisfying 
	\begin{equation*}
	\|z_1^*-z_2^*\|_\infty \le \beta\|x_1-x_2\|_2.
	\end{equation*}
	By the Cauchy-Schwarz inequality, this yields that
	\begin{equation*}
	|c^\top(z_1^*-z_2^*)| \le \|c\|_1 \|z_1^*-z_2^*\|_\infty \le \beta\|c\|_1\|x_1-x_2\|_2.
	\end{equation*}
	Defining $L=\beta\|c\|_1$ completes the proof.
\end{proof}

Lemma \ref{lem: lipschitz_lp_relaxations} shows that a partitioned LP relaxation is a Lipschitz relaxation over any chosen input part. This property will be used to derive a bound on the difference between the partitioned LP relaxation and the unrelaxed problem for one-layer networks based on the diameters of the input parts.

\begin{definition}[Diameter]
	\label{def: diameter}
	For a set $\mathcal{X}\subseteq\mathbb{R}^n$, the \emph{diameter} of $\mathcal{X}$ is defined as $d(\mathcal{X})=\sup\{ \|x-y\|_2 : x,y\in\mathcal{X} \}$.
\end{definition}

\begin{proposition}[Diameter bound]
	\label{prop: diameter_bound}
	Consider a one-layer feedforward neural network over the input uncertainty set $\mathcal{X}$ and the relaxed network constraint set $\mathcal{N}$. Let $\{\mathcal{X}^{(j)} \subseteq \mathcal{X} : j\in\{1,2,\dots,p\}\}$ be a partition of $\mathcal{X}$. Denote the largest diameter among the input parts by $d^*$, i.e., $d^* = \max\{d(\mathcal{X}^{(j)}) : j\in\{1,2,\dots,p\}\}$. Then, there exists a finite constant $L\in\mathbb{R}$ such that
	\begin{equation}
	\left|f^*(\mathcal{X}) - \max_{j\in\{1,2,\dots,p\}} \hat{f}^*(\mathcal{X}^{(j)})\right|\le Ld^*.
	\label{eq: diameter_bound}
	\end{equation}
\end{proposition}

\begin{proof}
	First, let $j^*$ be the index corresponding to the partition subproblem with the highest objective value: $\hat{f}^*(\mathcal{X}^{(j^*)})=\max_{j\in\{1,2,\dots,p\}} \hat{f}^*(\mathcal{X}^{(j)})$. By Lemma \ref{lem: lipschitz_lp_relaxations}, the network has an $L$-Lipschitz relaxation with respect to $\mathcal{N}^{(j^*)}$ on $\mathcal{X}^{(j^*)}$. Thus, there exists a finite constant $L\in\mathbb{R}$ such that
	\begin{equation*}
	L \ge \frac{|c^\top z_1^* - c^\top z_2^*|}{\|x_1-x_2\|_2}
	\end{equation*}
	for all $x_1,x_2\in\mathcal{X}^{(j^*)}$, where $c^\top z_1^* = \sup\{c^\top z : (x_1,z)\in\mathcal{N}^{(j^*)}\}$ and $c^\top z_2^*=\sup\{c^\top z : (x_2,z)\in\mathcal{N}^{(j^*)}\}$. Furthermore, by the definition of $d^*$ and $j^*$, we have that
	\begin{equation*}
	d^* \ge d(\mathcal{X}^{(j^*)}) \ge \|x_1-x_2\|_2 ~ \text{for all $x_1,x_2\in\mathcal{X}^{(j^*)}$}.
	\end{equation*}
	Let $\bar{x}\in\mathcal{X}^{(j^*)}$ be such that $W\bar{x} = l^{(j^*)}$ and $\bar{z} = \relu(W\bar{x})$, so that $c^\top \bar{z} = \sup\{c^\top z : (\bar{x},z)\in\mathcal{N}^{(j^*)}\}$. Furthermore, let $(\tilde{x}^*,\tilde{z}^*)$ denote a solution to the relaxed problem (\ref{eq: lp_relaxation}) over the $j^{*\text{th}}$ input part, i.e., corresponding to $\hat{f}^*(\mathcal{X}^{(j^*)})$. Then, since $\bar{x},\tilde{x}^*\in\mathcal{X}^{(j^*)}$ and $c^\top \tilde{z}^* = \sup\{c^\top z : (\tilde{x}^*,z)\in\mathcal{N}^{(j^*)}\}$, it holds that
	\begin{equation*}
	Ld^* \ge \frac{|c^\top \bar{z} - c^\top \tilde{z}^*|}{\|\bar{x}-\tilde{x}^*\|_2}\|\bar{x}-\tilde{x}^*\|_2 = |c^\top \bar{z}-\hat{f}^*(\mathcal{X}^{(j^*)})|.
	\end{equation*}
	Now, since $(\bar{x},\bar{z})$ is feasible for the unrelaxed problem (\ref{eq: robustness_certification_problem}) and the relaxation $\hat{f}^*(\mathcal{X}^{(j^*)})$ provides an upper bound on the unrelaxed problem by Proposition \ref{prop: partitioned_relaxation_bound}, it holds that
	\begin{equation*}
	c^\top \bar{z} \le f^*(\mathcal{X}) \le \hat{f}^*(\mathcal{X}^{(j^*)}).
	\end{equation*}
	This implies that $|f^*(\mathcal{X}) - \hat{f}^*(\mathcal{X}^{(j^*)})| \le |c^\top\bar{z} - \hat{f}^*(\mathcal{X}^{(j^*)})|$. Therefore, $Ld^* \ge |f^*(\mathcal{X}) - \hat{f}^*(\mathcal{X}^{(j^*)})|$, proving \eqref{eq: diameter_bound}, as desired.
\end{proof}

In the case that the network has a Lipschitz relaxation that is uniform over all possible input parts, Proposition \ref{prop: diameter_bound} shows that as the partition becomes finer, namely $d^*\to 0$, the gap between the partitioned relaxation and the true solution converges to zero. As a result, partitioned relaxations are asymptotically exact.

\section{Optimal Partitioning}
\label{sec: optimal_partitioning}
In this section, we construct a partition with a finite number of input parts under which LP relaxations exactly recover the nonconvex robustness certification problem. Motivated by this optimal partition, we develop a simple and computationally tractable partitioning procedure that minimizes the worst-case relaxation error.

\subsection{Exact Partitioned Relaxation}
The goal is to show that by meticulously selecting the partition of the input uncertainty set based on the rows of the weight matrix $W$, the relaxation introduced by the resulting linear programs becomes exact.

\begin{proposition}[Motivating partition]
	\label{prop: motivating_partition}
	Consider a one-layer feedforward neural network and denote the $i^{\text{th}}$ row of $W$ by $w_i^\top\in\mathbb{R}^{1\times n_x}$ for all $i\in\{1,2,\dots,n_z\}$. Define $\mathcal{J} = \{0,1\}^{n_z}$ and take the partition of $\mathcal{X}$ to be indexed by $\mathcal{J}$. That is, $\{\mathcal{X}^{(j)} \subseteq \mathcal{X} : j\in\mathcal{J}\}$, where for a given $j\in\mathcal{J}$ we define
	\begin{equation}
	\begin{aligned}
	\mathcal{X}^{(j)} ={}& \{x\in\mathcal{X} : w_i^\top x \ge 0 ~ \emph{for all $i$ such that $j_i=1$}, \\
	&{} w_i^\top x<0 ~ \emph{for all $i$ such that $j_i=0$}\}.
	\end{aligned} \label{eq: motivating_partition_input_part}
	\end{equation}
	Then, the partitioned relaxation is exact, i.e.,
	\begin{equation}
	f^*(\mathcal{X}) = \max_{j\in\mathcal{J}} \hat{f}^*(\mathcal{X}^{(j)}). \label{eq: motivating_partition_exact}
	\end{equation}
\end{proposition}

\begin{proof}
	We first show that $\{\mathcal{X}^{(j)} \subseteq \mathcal{X} : j\in\mathcal{J}\}$ is a valid partition. Since $\mathcal{X}^{(j)} \subseteq \mathcal{X}$ for all $j\in\mathcal{J}$, the relation $\cup_{j\in\mathcal{J}}\mathcal{X}^{(j)} \subseteq \mathcal{X}$ is satisfied. Now, suppose that $x\in\mathcal{X}$. Then, for all $i\in\{1,2,\dots,n_z\}$, it holds that either $w_i^\top x \ge 0$ or $w_i^\top x<0$. Define $j\in\{0,1\}^{n_z}$ as follows:
	\begin{equation*}
	j_i = \begin{aligned}
	\begin{cases}
	1 & \text{if $w_i^\top x \ge 0$}, \\
	0 & \text{if $w_i^\top x < 0$},
	\end{cases}
	\end{aligned}
	\end{equation*}
	for all $i\in\{1,2,\dots,n_z\}$. Then, by the definition of $\mathcal{X}^{(j)}$ in (\ref{eq: motivating_partition_input_part}), it holds that $x\in\mathcal{X}^{(j)}$. Therefore, the relation $x\in\mathcal{X}$ implies that $x\in\mathcal{X}^{(j)}$ for some $j\in\{0,1\}^{n_z} = \mathcal{J}$. Hence, $\mathcal{X}\subseteq \cup_{j\in\mathcal{J}}\mathcal{X}^{(j)}$, and therefore $\cup_{j\in\mathcal{J}}\mathcal{X}^{(j)} = \mathcal{X}$.
	
	We now show that $\mathcal{X}^{(j)}\cap\mathcal{X}^{(k)}=\emptyset$ for all $j\ne k$. Let $j,k\in\mathcal{J}$ with the property that $j\ne k$. Then there exists $i\in\{1,2,\dots,n_z\}$ such that $j_i\ne k_i$. Let $x\in\mathcal{X}^{(j)}$. In the case that $w_i^\top x\ge 0$, it holds that $j_i = 1$ and therefore $k_i = 0$. Hence, for all $y\in\mathcal{X}^{(k)}$, it holds that $w_i^\top y < 0$, and therefore $x\notin\mathcal{X}^{(k)}$. An analogous reasoning shows that $x\notin\mathcal{X}^{(k)}$ when $w_i^\top x < 0$. Therefore, one concludes that $x\in\mathcal{X}^{(j)}$ and $j\ne k$ implies that $x\notin\mathcal{X}^{(k)}$, i.e., that $\mathcal{X}^{(j)}\cap\mathcal{X}^{(k)} = \emptyset$. Hence, $\{\mathcal{X}^{(j)} \subseteq \mathcal{X} : j\in\mathcal{J}\}$ is a valid partition.
	
	We now prove (\ref{eq: motivating_partition_exact}). Let $j\in\mathcal{J}$. Since $w_i^\top x \ge 0$ for all $i$ such that $j_i=1$, the preactivation lower bound becomes $l_i^{(j)} = 0$ for all such $i$. On the other hand, since $w_i^\top x < 0$ for all $i$ such that $j_i=0$, the preactivation upper bound becomes $u_i^{(j)} = 0$ for all such $i$. Therefore, the relaxed network constraint set (\ref{eq: network_constraint_set}) for the $j^{\text{th}}$ input part reduces to
	\begin{align*}
	\mathcal{N}^{(j)} ={}& \{(x,z)\in\mathbb{R}^{n_x}\times\mathbb{R}^{n_z} : \\
	&{} z_i = 0 ~ \text{for all $i$ such that $j_i = 0$}, \\
	&{} z_i = w_i^\top x = (Wx)_i ~ \text{for all $i$ such that $j_i = 1$}\}.
	\end{align*}
	That is, the relaxed ReLU constraint envelope collapses to the exact ReLU constraint through the \textit{a priori} knowledge of each preactivation coordinate's sign.
%
%
	Therefore, we find that for all $x\in\mathcal{X}^{(j)}$ it holds that $(x,z)\in\mathcal{N}^{(j)}$ if and only if $z=\relu(Wx)$. Hence, the LP over the $j^{\text{th}}$ input part yields that
	\begin{align*}
	\hat{f}^*(\mathcal{X}^{(j)}) ={}& \sup\{c^\top z : (x,z)\in\mathcal{N}^{(j)}, ~ x\in\mathcal{X}^{(j)}\} \\
	={}& \sup\{c^\top z : z = \relu(Wx), ~ x\in\mathcal{X}^{(j)}\} \\
	\le{}& \sup\{c^\top z : z = \relu(Wx), ~ x\in\mathcal{X}\} \\
	={}& f^*(\mathcal{X}).
	\end{align*}
	Therefore, $\max_{j\in\mathcal{J}}\hat{f}^*(\mathcal{X}^{(j)}) \le f^*(\mathcal{X})$, since $j$ was chosen arbitrarily. Since $f^*(\mathcal{X})\le \max_{j\in\mathcal{J}}\hat{f}^*(\mathcal{X}^{(j)})$ by the relaxation bound (\ref{eq: partitioned_relaxation_bound}), the equality (\ref{eq: motivating_partition_exact}) holds, as desired.
\end{proof}

\begin{figure}[ht]
    \vspace*{-0.5\baselineskip}
    \centering
    \includegraphics[width=0.9\linewidth]{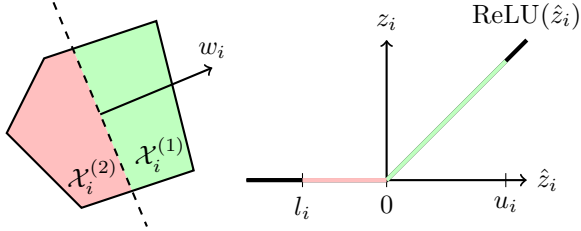}
    \caption{Partitioning based on row $w_i^\top$ of the weight matrix. This partition results in an exact ReLU constraint in coordinate $i$ over the two resulting input parts $\mathcal{X}_i^{(1)} = \{x\in\mathcal{X} : w_i^\top x \ge 0\}$ and $\mathcal{X}_i^{(2)}=\mathcal{X}\setminus\mathcal{X}_i^{(1)}$.}
    \label{fig: weight-based_partition}
    \vspace*{-\baselineskip}
\end{figure}

The partition introduced in Proposition \ref{prop: motivating_partition} requires solving $2^{n_z}$ linear programs, which may quickly become computationally intractable in practice. Despite this limitation, the result provides two major theoretical implications. First, it shows that, using the input partitioning methodology presented in this paper, the robustness certification problem can be solved exactly via a finite number of linear program subproblems. Second, Proposition \ref{prop: motivating_partition} provides a starting point to answer the following question: If the input uncertainty set is to be partitioned into only two parts, what is the optimal partition to choose? The proposition gives insight into the structure of an optimal partition, namely that it is defined by intersections of the half-spaces generated by the rows of $W$ (see Fig.\ \ref{fig: weight-based_partition}). Motivated by this structure, we develop an optimal two-part partitioning scheme in the next section.

\subsection{Optimal Partitioning Scheme}
To derive an optimal partitioning scheme, we first bound the relaxation error in the worst-case sense.

\begin{theorem}[Worst-case relaxation bound]
	\label{thm: worst-case_relaxation_bound}
	Consider a one-layer feedforward neural network with the input uncertainty set $\mathcal{X}$ and preactivation bounds $l,u\in\mathbb{R}^{n_z}$. Consider also the relaxation error $\Delta f^*(\mathcal{X}) \coloneqq \hat{f}^*(\mathcal{X})-f^*(\mathcal{X})$. Assume that there exists $x^*\in\mathcal{X}$ such that $(x^*,\tilde{z}^*)$ and $(x^*,z^*)$ are optimal solutions for the relaxation $\hat{f}^*(\mathcal{X})$ and the unrelaxed problem $f^*(\mathcal{X})$, respectively. Then, it holds that
	\begin{equation}
	\Delta f^*(\mathcal{X}) \le -\sum_{i=1}^{n_z} \relu(c_i)\frac{u_i l_i}{u_i-l_i}. \label{eq: worst-case_relaxation_bound}
	\end{equation}
\end{theorem}

\begin{proof}
    The definitions of $(x^*,\tilde{z}^*)$ and $(x^*,z^*)$ give that
	\begin{equation}
	\Delta f^*(\mathcal{X}) = \sum_{i=1}^{n_z}c_i(\tilde{z}^*_i-z^*_i) \le \sum_{i=1}^{n_z}\Delta f^*_i, \label{eq: worst-case_relaxation_bound_intermediate_1}
	\end{equation}
	where
	\begin{align*}
	\Delta f^*_i ={}& \sup\{c_i(\tilde{z}_i-z_i) : z_i = \relu(w_i^\top x), ~ \tilde{z}_i\ge 0, \\
	&{} \tilde{z}_i \ge w_i^\top x, ~ \tilde{z}_i \le \frac{u_i}{u_i-l_i}(w_i^\top x - l_i), ~ x\in\mathcal{X} \}
	\end{align*}
	for all $i\in\{1,2,\dots,n_z\}$. For $c_i\ge 0$, it holds that
	\begin{align*}
	\Delta f_i^* ={}& c_i\sup\{\tilde{z}_i-z_i : z_i = \relu(w_i^\top x), ~ \tilde{z}_i\ge 0, \\
	&{} \tilde{z}_i \ge w_i^\top x, ~ \tilde{z}_i \le \frac{u_i}{u_i-l_i}(w_i^\top x - l_i), ~ x\in\mathcal{X} \} \\
	={}& -c_i\frac{u_i l_i}{u_i-l_i}, 
	\end{align*}
	where the final equality is readily shown by computing the maximum difference between the line $\tilde{z}_i = \frac{u_i}{u_i-l_i}(\hat{z}_i-l_i)$ and the function $z_i=\relu(\hat{z}_i)$ over $\hat{z}_i\in[l_i,u_i]$. On the other hand, for $c_i<0$, we have that
	\begin{align*}
	\Delta f_i^* ={}& c_i\inf\{\tilde{z}_i-z_i : z_i = \relu(w_i^\top x), ~ \tilde{z}_i\ge 0, \\
	&{} \tilde{z}_i \ge w_i^\top x, ~ \tilde{z}_i \le \frac{u_i}{u_i-l_i}(w_i^\top x - l_i), ~ x\in\mathcal{X}\} \\
	={}& 0,
	\end{align*}
	where the final equality is due to the fact that $\tilde{z}_i\ge z_i$ on the above feasible set and $\tilde{z}_i=z_i=0$ is feasible. Substituting these results into (\ref{eq: worst-case_relaxation_bound_intermediate_1}) gives the desired bound (\ref{eq: worst-case_relaxation_bound}).
\end{proof}

The value $\Delta f^*_i$ can be interpreted as the worst-case relaxation error in coordinate $i$. From this perspective, Theorem \ref{thm: worst-case_relaxation_bound} provides an upper bound on the overall worst-case relaxation error. The error bound scales linearly as the input uncertainty set is made smaller, as shown in Corollary \ref{cor: linear_scaling_of_relaxation_error} that follows.

\begin{corollary}[Linear scaling of relaxation error]
    \label{cor: linear_scaling_of_relaxation_error}
    Under the settings of Theorem \ref{thm: worst-case_relaxation_bound}, consider an input uncertainty subset $\tilde{\mathcal{X}}\subseteq \mathcal{X}$ such that its associated preactivation bounds are scaled inward by a factor of $\alpha\in(0,1)$, namely $\tilde{u} = \alpha u$ and $\tilde{l} = \alpha l$. Then, the worst-case relaxation bound over $\tilde{\mathcal{X}}$ also scales by $\alpha$, i.e.,
    \begin{equation}
        \hat{f}^*(\tilde{\mathcal{X}}) - f^*(\mathcal{X}) \le -\alpha\sum_{i=1}^{n_z} \relu(c_i) \frac{u_i l_i}{u_i - l_i}. \label{eq: linear_scaling_of_relaxation_error}
    \end{equation}
\end{corollary}

\begin{proof}
    Since $\tilde{\mathcal{X}}\subseteq\mathcal{X}$, it holds that $f^*(\tilde{\mathcal{X}})\le f^*(\mathcal{X})$. Therefore, by Theorem \ref{thm: worst-case_relaxation_bound} on $\tilde{\mathcal{X}}$ we have that
    \begin{equation*}
        \begin{aligned}
        \hat{f}^*(\tilde{\mathcal{X}}) - f^*(\mathcal{X}) \le{}& \hat{f}^*(\tilde{\mathcal{X}}) - f^*(\tilde{\mathcal{X}}) \\
        \le{}& - \sum_{i=1}^{n_z} \relu(c_i) \frac{\tilde{u}_i \tilde{l}_i}{\tilde{u}_i-\tilde{l}_i}.
        \end{aligned}
    \end{equation*}
    Substituting $\tilde{u}_i=\alpha u_i$ and $\tilde{l}_i = \alpha l_i$ completes the proof.
\end{proof}

We now focus on developing an optimal two-part partitioning scheme based on the worst-case relaxation bound of Theorem \ref{thm: worst-case_relaxation_bound}. We start with the following lemma.

\begin{lemma}[Two-part partition bound]
    \label{lem: two-part_partition_bound}
    Under the settings of Theorem \ref{thm: worst-case_relaxation_bound}, let $i\in\{1,2,\dots,n_z\}$ and consider a two-part partition of $\mathcal{X}$ given by $\{\mathcal{X}_i^{(1)},\mathcal{X}_i^{(2)}\}$, where $\mathcal{X}_i^{(1)} = \{x\in\mathcal{X} : w_i^\top x \ge 0\}$ and $\mathcal{X}_i^{(2)} = \mathcal{X}\setminus \mathcal{X}_i^{(1)}$. Consider also the partitioned relaxation error $\Delta f^*(\{\mathcal{X}_i^{(1)},\mathcal{X}_i^{(2)}\}) \coloneqq \max_{j\in\{1,2\}} \hat{f}^*(\mathcal{X}_i^{(j)}) - f^*(\mathcal{X})$. It holds that
    \begin{equation}
        \Delta f^*(\{\mathcal{X}_i^{(1)},\mathcal{X}_i^{(2)}\}) \le -\sum_{\substack{k=1 \\ k\ne i}}^{n_z} \relu(c_k) \frac{u_k l_k}{u_k - l_k}. \label{eq: two-part_partition_bound}
    \end{equation}
\end{lemma}

\begin{proof}
    Consider the relaxation solved over the first input part, $\mathcal{X}_i^{(1)}$, and denote by $l^{(1)},u^{(1)}\in\mathbb{R}^{n_z}$ the corresponding preactivation bounds. Since $w_i^\top x \ge 0$ on this input part, the preactivation bounds for the first subproblem $\hat{f}^*(\mathcal{X}_i^{(1)})$ can be taken as $l^{(1)}=(l_1,l_2,\dots,l_{i-1},0,l_{i+1},\dots,l_{n_z})$ and $u^{(1)}=u$. It follows from Theorem \ref{thm: worst-case_relaxation_bound} that
    \begin{equation*}
        \begin{aligned}
        \hat{f}^*(\mathcal{X}_i^{(1)}) - f^*(\mathcal{X}_i^{(1)}) \le{}& -\sum_{k=1}^{n_z}\relu(c_k)\frac{u_k^{(1)}l_k^{(1)}}{u_k^{(1)}-l_k^{(1)}} \\
        ={}& -\sum_{\substack{k=1 \\ k\ne i}}^{n_z}\relu(c_k)\frac{u_k l_k}{u_k-l_k}. 
        \end{aligned}
    \end{equation*}
    Similarly, over the second input part, $\mathcal{X}_i^{(2)}$, we have that $w_i^\top x < 0$, and so the preactivation bounds for the second subproblem $\hat{f}^*(\mathcal{X}_i^{(2)})$ can be taken as $l^{(2)}=l$ and $u^{(2)} = (u_1,u_2,\dots,u_{i-1},0,u_{i+1},\dots,u_{n_z})$, resulting in the same bound: $\hat{f}^*(\mathcal{X}_i^{(2)}) - f^*(\mathcal{X}_i^{(2)}) \le -\sum_{\substack{k=1 \\ k\ne i}}^{n_z}\relu(c_k)\frac{u_k l_k}{u_k-l_k}$. Putting these two bounds together and using the fact that $f^*(\mathcal{X}_i^{(j)})\le f^*(\mathcal{X})$ for all $j\in\{1,2\}$, we find that
    \begin{multline*}
        \begin{aligned}
        \Delta f^*(\{\mathcal{X}_i^{(1)},\mathcal{X}_i^{(2)}\}) ={}& \max_{j\in\{1,2\}} \left( \hat{f}^*(\mathcal{X}_i^{(j)}) - f^*(\mathcal{X}) \right) \\
        \le{}& \max_{j\in\{1,2\}}\left( \hat{f}^*(\mathcal{X}_i^{(j)}) - f^*(\mathcal{X}_i^{(j)}) \right) \\
        \le{}& -\sum_{\substack{k=1 \\ k\ne i}}^{n_z}\relu(c_k)\frac{u_k l_k}{u_k-l_k},
        \end{aligned}
    \end{multline*}
    as desired.
\end{proof}

With the partitioned relaxation error bound of Lemma \ref{lem: two-part_partition_bound} established, we now present the optimal two-part partition.

\begin{theorem}[Optimal partition]
	\label{thm: optimal_partition}
	Consider the two-part partitions defined by the rows of $W$: $\{\mathcal{X}_i^{(1)},\mathcal{X}_i^{(2)}\}$, where $\mathcal{X}_i^{(1)} = \{ x\in\mathcal{X} : w_i^\top x \ge 0 \}$ and $\mathcal{X}_i^{(2)} = \mathcal{X}\setminus \mathcal{X}_i^{(1)}$ for all $i\in \{1,2,\dots,n_z\} \eqqcolon \mathcal{I}$. The optimal partition that minimizes the worst-case relaxation error in (\ref{eq: two-part_partition_bound}) is given by
	\begin{equation}
	i^* \in \arg\min_{i\in\mathcal{I}} \relu(c_i)\frac{u_i l_i}{u_i-l_i}. \label{eq: optimal_partition}
	\end{equation}
\end{theorem}

\begin{proof}
	Minimizing the bound in (\ref{eq: two-part_partition_bound}) of Lemma \ref{lem: two-part_partition_bound} over the partition $i$ gives rise to
	\begin{multline*}
	\min_{i\in\mathcal{I}} \left(-\sum_{\substack{k=1\\k\ne i}}^{n_z}\relu(c_k)\frac{u_k l_k}{u_k - l_k}\right) \\
	= -\sum_{k=1}^{n_z}\relu(c_k)\frac{u_k l_k}{u_k - l_k} + \min_{i\in\mathcal{I}}\relu(c_i)\frac{u_i l_i}{u_i - l_i},
	\end{multline*}
	as desired.
\end{proof}

Theorem \ref{thm: optimal_partition} provides a methodical way of selecting the optimal two-part partition based on the rows of $W$ in a worst-case sense. To understand the efficiency of this result, notice that the optimization over $i$ scales linearly with the dimension $n_z$, and the resulting LP subproblems require the addition of only one extra linear constraint. Finally, we note that Theorem \ref{thm: optimal_partition} can be immediately extended in two ways. First, by ordering the values $\relu(c_i)\frac{u_i l_i}{u_i-l_i}$ in \eqref{eq: optimal_partition}, we can choose the best $n_p>1$ rows to partition along in order to perform a $2^{n_p}$-part partition. The other application of Theorem \ref{thm: optimal_partition} is the following recursion: solve the two-part partitioned LP using Theorem \ref{thm: optimal_partition} to partition $\mathcal{X}$. If $\mathcal{X}_{i^*}^{(j^*)}$ is the input part containing the solution $x^*$, then partition $\mathcal{X}_{i^*}^{(j^*)}$ into two smaller sub-parts again according to Theorem \ref{thm: optimal_partition} and solve the refined partitioned LP over the sub-parts. Continuing this nested procedure results in tightened localization of a true worst-case input (i.e., a solution to $f^*(\mathcal{X})$) and further reduces conservatism of the partitioned LP relaxation.

\section{Simulation Results}
\label{sec: simulation_results}
Consider a one-layer classification network with four inputs and three outputs, trained on the celebrated Iris data set \cite{fisher1936use} with a test accuracy of $97\%$. A negative optimal objective value in the robustness certification problem indicates that any perturbation in $\mathcal{X} = \{x\in\mathbb{R}^{n_x}:\|x-\bar{x}\|_\infty \le \epsilon\}$ of the nominal input $\bar{x}$ will not change the input's classification. For this experiment, we solve the certification problem for 10 different nominal inputs using \textsc{Matlab} and CVX on a Windows 7 laptop with a 2.9 GHz quad-core i7 processor.

We first solve the problem in its nonconvex form (using multistart and \textsc{Matlab}'s \texttt{fmincon} function). We then solve the problem using an unpartitioned LP relaxation, and then using partitioned LP relaxations, one per row of $W$. We restrict our experiments to two-part partitions to explore the effect of changing the row of $W$ along which we partition. The average time taken to solve the nonconvex, unpartitioned LP, and partitioned LPs are $0.21$, $0.26$, and $0.48$ seconds, respectively. As expected, the computational burden of the two-part partitioned LP is twice that of the unpartitioned LP, both of which are very fast for this network.

\begin{figure}[ht]
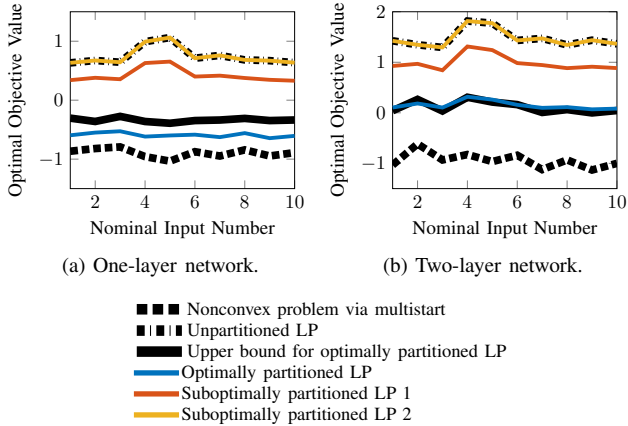

    \vspace*{-\baselineskip}
    \setlength\fwidth{0.475\linewidth}
    \centering
    \subfloat[One-layer network.]{%
		\centering
		\includegraphics[width=\fwidth]{figures/partitioned_objectives_iris.tikz}
		\label{fig: partitioned_objectives_iris}
	}
	\hfil
	\subfloat[Two-layer network.]{%
		\centering
		\includegraphics[width=\fwidth]{figures/partitioned_objectives_iris_multilayer.tikz}
		\label{fig: partitioned_objectives_iris_multilayer}
	} \\
	{\scriptsize
	\vspace*{\baselineskip}
	$\begin{array}{l}
	\ref{plt: line_1} \text{Nonconvex problem via multistart} \\
	\ref{plt: line_2} \text{Unpartitioned LP} \\
	\ref{plt: line_3} \text{Upper bound for optimally partitioned LP} \\
	\ref{plt: line_4} \text{Optimally partitioned LP} \\
	\ref{plt: line_5} \text{Suboptimally partitioned LP 1} \\
	\ref{plt: line_6} \text{Suboptimally partitioned LP 2}
	\end{array}$
	}
    \caption{Optimal values of robustness certification for ReLU Iris classifier. For the two-layer network, the optimal partitioning scheme is applied to only the ReLU constraints of the final layer.}
    \label{fig: iris_experiment}
    \vspace*{-\baselineskip}
\end{figure}

The optimal objective values for each nominal input are shown in Fig.\ \ref{fig: partitioned_objectives_iris}. As seen, the optimally partitioned LP developed in Theorem \ref{thm: optimal_partition} yields the best convex upper bound on the true problem. Furthermore, ordering the rows $w_i^\top$ by their suboptimality in (\ref{eq: optimal_partition}) corresponds to the order of relaxation tightening. For instance, in Fig.\ \ref{fig: iris_experiment}, we compare the LP partitioned via $i_1 \in \arg\min_{i\in\mathcal{I}\setminus \{i^*\}} \relu(c_i)\frac{u_il_i}{u_i-l_i}$ (suboptimally partitioned LP 1) and that partitioned via $i_2 \in \arg\min_{i\in \mathcal{I}\setminus\{i^*,i_1\}}\relu(c_i)\frac{u_il_i}{u_i-l_i}$ (suboptimally partitioned LP 2). In this example, the suboptimally partitioned LP 2 (partitioned along worst row $w_{i_2}^\top$) coincides with the unpartitioned LP, suggesting that none of the relaxation error is attributed to the $i_2^{\text{th}}$ coordinate of the ReLU layer. The suboptimally partitioned LPs do not certify robustness, as the objective values are positive for each nominal input tested. On the other hand, the developed optimal partitioning scheme tightens the relaxation enough to provide a certificate of robustness for the one-layer network corresponding to every nominal input tested here. For the same experiment on a two-layer network (with an added five-neuron ReLU layer), we find that the optimal partitioning scheme maintains the best convex upper bound, albeit without guaranteed relaxation error bounds (see Fig.\ \ref{fig: partitioned_objectives_iris_multilayer}). The average computation times for the nonconvex, unpartitioned LP, and partitioned LPs rise to $0.94$, $0.68$, and $1.48$ seconds, respectively. For general network sizes, the two-part partitioned LP maintains the polynomial-time complexity of linear programming with respect to the number of neurons, since it requires solving two instances of the same LP structure \cite{karmarkar1984new}.

\section{Conclusions}
\label{sec: conclusions}
In this work, we develop a partition-based method for ReLU neural network robustness certification that systematically reduces relaxation error while maintaining the efficiency of linear programming. We theoretically justify the effectiveness of partitioning and derive an optimal partitioning scheme. A case study on real data shows that the proposed method is able to certify the robustness of a network while the existing methods fail. Our results demonstrate, both theoretically and experimentally, that partition-based certification procedures are capable of tightening relaxation errors with remarkable simplicity.


\bibliographystyle{IEEEtran}
\bibliography{references}


\end{document}